\documentclass{article}

\usepackage{multicol} 
\usepackage[svgnames]{xcolor} 

\usepackage{times} 

\usepackage{graphicx} 
\graphicspath{{figures/}} 
\usepackage{booktabs} 
\usepackage[font=small,labelfont=bf]{caption} 
\usepackage{amsfonts, amsmath, amsthm, amssymb, bm} 
\usepackage{wrapfig} 

\usepackage{mathtools} 
\DeclarePairedDelimiterX{\norm}[1]{\lVert}{\rVert}{#1}

\usepackage[utf8]{inputenc} 
\usepackage[T1]{fontenc}    
\usepackage{hyperref}       
\usepackage{url}            
\usepackage{booktabs}       
\usepackage{amsfonts}       
\usepackage{microtype}      

\usepackage{algorithm}
\usepackage{algorithmic}

\usepackage{natbib}
\usepackage{pifont}
\usepackage{xcolor}

\title{Neograd: Near-Ideal Gradient Descent}

\author{
  Michael F.~Zimmer \\
  \texttt{zim@neomath.com} \\
}

\begin{document}

\maketitle

\begin{abstract}
The purpose of this paper is to improve upon existing variants of gradient descent by solving two problems: 
(1) removing (or reducing) the plateau that occurs while minimizing the cost function,
(2) continually adjusting the learning rate to an "ideal" value.
The approach taken is to approximately solve for the learning rate as a function of a trust metric.
When this technique is hybridized with momentum, it creates an especially effective gradient descent variant, called NeogradM.
It is shown to outperform Adam on several test problems, and can easily reach cost function values that are smaller by a factor of $10^8$, for example.
\end{abstract}

\section{Introduction}

Gradient descent \citep{Press-book2007} is an iterative optimization algorithm for differentiable functions.  Given a cost function (CF) $f$ which depends on the parameter vector
${\bm \theta} = (\theta_1, \theta_2, ...)$, the initial choice of ${\bm \theta}$ is updated as
\begin{align*}
{\bm \theta}_{new} & = {\bm \theta}_{old} - \alpha {\bm \nabla} f ,
\end{align*}
where ${\bm \nabla} f$ is the gradient of $f$ with respect to ${\bm \theta}$, and $\alpha$ is a positive scalar known as the learning rate.  
While the origin of this method dates back to \cite{Cauchy1847}
\footnote{For some discussion, see \cite{Lemarechal2010}.}
, and while it has seen use in a variety of disciplines, it has found focused attention in the field of machine learning (ML).

It is also the case that despite its widespread use with many improvements introduced over the decades, the best variants
still have issues.  Perhaps the two most prominent are: (1) avoiding a plateau in the CF vs iteration graph;
(2) efficiently determining a good value for $\alpha$.
These are the problems which this paper seeks to address.  
A metric will be introduced which provides an assessment of the progress of the GD updates.  
Following that, a class of algorithms is introduced (Neograd), which is based on keeping this metric at a constant value.
This controlled approach to optimization will be shown to lead to superior results in the examples discussed.

In the following section, {\em Related Work}, a number of methods used in comparable iterative algorithms are reviewed.
This is followed by the section {\em The Diagnostic Metric $\rho$}, which introduces the metric which measures the validity of a linear approximation.
The next section, {\em Algorithms}, demonstrates the utility of the {\em constant $\rho$ ansatz}, in the Ideal and Near-Ideal cases.
Also, it is here that the Neograd family of algorithms is defined.
Next, in the {\em Experiments} section, three main examples are covered: (1) sigmoid-well CF which demonstrates the adaptivity of $\alpha$;
(2) Beale's function, which highlights some of the differences between Adam and NeogradM, and also shows NeogradM's superior performance;
(3) a cross-entropy CF used in the context of a digit recognition problem, which also highlights the performance of NeogradM.
In addition, Appendices A and B provide pseudocode for the Neograd algorithms, and Appendix C provides a demonstration of the stability of the new adaptation formula introduced herein.

\section{Related Work}
\label{sec:related}

There has been a great deal of research on iterative approaches to finding a local minimum of a function
(e.g., see \citep{Nocedal-book2006}, \citep{Nesterov-book2018}, \citep{Sra-book2011}).
They can largely be classified as either line search and trust region approaches.
Other optimization themes such as momentum, stochastic gradient descent, and other approaches will also be discussed.
The primary focus here will be on first-order methods
\footnote{For a discussion of the issues with second-order methods, see section 5.4 in \cite{Bishop-book2006}.}
such as GD, since they have much lower complexity costs per iteration than second-order methods in high-dimensional parameter spaces. 

With \underline{Line Search}, the basic idea is to first determine a direction (${\bm p}_n$) and then a step size ($\alpha_n$) for reducing the function, from a value $f({\bm \theta}_n )$ to $f({\bm \theta}_n + \alpha {\bm p}_n )$, for the $n$th iteration.  The direction ${\bm p}_n$ may be written as
${\bm p}_n = - {\bm B}^{-1} {\bm \nabla} f$
where in the case of gradient descent ${\bm B}$ is the identity matrix, and in the case of Newton's method it is the Hessian of $f$.
Typically, $\alpha_n$ is chosen to minimize $f({\bm \theta}_n + \alpha_n {\bm p}_n )$, and a number of trial $\alpha_n$ values are considered toward that end.
Also, $\alpha_n$ may be chosen to satisfy these conditions
\begin{subequations}
\begin{align}
f({\bm \theta}_n + \alpha_n {\bm p}_n ) & \leq f({\bm \theta}_n) + c_1 \alpha_n {\bm \nabla f}_n^T {\bm p}_n   \\
{\bm \nabla} f({\bm \theta}_n + \alpha_n {\bm p})^T {\bm p}_n & \geq c_2 {\bm \nabla} f_n^T {\bm p}_n ,
\end{align}
\label{eqn:Wolfe}
\end{subequations}
where $0 < c_1 < c_2 < 1$.
The first of these conditions, known as the Armijo condition \citep{Armijo1966},
stipulates that an $\alpha_n$ should lead to a {\em sufficient decrease} in $f$;
the second stipulates that very small updates are not taken.
Together, they are known as the Wolfe conditions \citep{Wolfe1969,Wolfe1971}.
Alternatively, one could use the Barzilai-Borwein method for setting the learning rate \citep{Barzilai1988}, via
\begin{equation*}
\alpha_n = \frac{ | ({\bm \theta}_n - {\bm \theta}_{n-1})^T ( {\bm \nabla} f( {\bm \theta}_n) - {\bm \nabla} f( {\bm \theta}_{n-1}) ) |  }
{  \| {\bm \nabla} f( {\bm \theta}_n) - {\bm \nabla} f( {\bm \theta}_{n-1})  \| } .
\end{equation*}

\underline{Trust Region} methods begin by modeling a function with a quadratic model
\begin{align*}
m_n ( {\bm p}_n ) & = f_n + {\bm \nabla}f_n^T {\bm p}_n + \frac{1}{2} {\bm p}^T_n {\bm B}_n {\bm p}_n ,
\end{align*}
where ${\bm B}$ is a symmetric matrix that is meant to approximate the Hessian of $f$.
A solution is sought to $\min_p m_n(p)$ for $\|p \| \leq \Delta$, where $\Delta$ is a radius defining the trust region.
Whether $m_n$ is a good fit is decided during the iterations by examining the ratio
\begin{align}
r_n & = \frac{ f( {\bm \theta}_n ) - f( {\bm \theta}_n + {\bm p}_n ) }{ m( {\bm 0} ) - m( {\bm p}_n ) } .
\label{eqn:trustmetric}
\end{align}
The numerator is the {\em actual reduction} of $f$, and the denominator is the {\em predicted reduction}.
The radius $\Delta$ is adjusted in order to (ideally) keep this metric in the range $0.25-0.5$.  
When $r_n < 0.25$ it indicates the step size is too large, and $r_n > 0.50$ it is too small.
Although this is a second-order method, it is of interest here because of the general methodology and the ratio $r_n$,
which will be compared to the metric introduced in this paper.

For at least basic GD, and probably most other gradient-based optimization, it's the case that the updates
would benefit from being smoothed out.
For example, it is often the case that basic GD will lead to oscillations \citep{Press-book2007}, as it continually overshoots a minimum.
A smoothing effect can be achieved by introducing \underline{Momentum}, which can be implemented by an exponential weighted average of the gradient, as shown here with the variable ${\bm v}$ \citep{Polyak1964}
\begin{align*}
{\bm v}_n & = \beta {\bm v}_{n-1} + \alpha {\bm \nabla} f ( {\bm \theta}_{n-1} ) \\
{\bm \theta}_n & = {\bm \theta}_{n-1} - {\bm v}_n ,
\end{align*}
where the momentum parameter $\beta$ is a nonnegative number less than 1. 
Nesterov accelerated gradient \citep{Nesterov1983,Nesterov-book2018} is another version of momentum which has been noted to achieve better results; in its original form it requires an additional,  {\em look-ahead} evaluation of the gradient
\begin{align*}
{\bm v}_n & = \beta {\bm v}_{n-1} + \alpha {\bm \nabla} f ( {\bm \theta}_{n-1} - \beta {\bm v}_{n-1} ) \\
{\bm \theta}_n & = {\bm \theta}_{n-1} - {\bm v}_n .
\end{align*}

\underline{Stochastic Gradient Descent (SGD)} is a departure from the above methods, in that 
rather than using a single data set (and hence a single CF), different samples of the data set are used for each update.
It follows that the CF being minimized changes with each iteration, although they are presumably statistically related since they're drawn
from the same underlying distribution.
There have been a series of algorithms introduced which allow for different effective learning rates for each component $\theta_i$.
The concepts in play were momentum applied to the gradient and possibly the gradient squared.
For this category, one may group together the approaches of AdaGrad \citep{Duchi2011}, AdaDelta \citep{Zeiler2012}, RMS Prop \citep{Tieleman2012}, Adam \& AdaMax \citep{Kingma2015},
Nadam \citep{Dozat2016}, AdamNC \citep{Reddi2018},
and others \citep[see][] {Ruder2017}.
At this point, Adam seems to be the preferred choice, and is the one used for comparisons in this paper.
Its innovation was to additionally use momentum in ${\bm \nabla} f$ within RMSProp.
The main updates
\footnote{Additionally, the variables ${\bm m}_n$ and ${\bm v}_n$ are corrected for bias at each step.
The operations in ${\bm g}_n^2$ and ${\bm m}_n / (\sqrt{ {\bm v}_n } + \epsilon )$ are understood to be performed element-wise.}
of \underline{Adam} are
\begin{align*}
{\bm m}_n & = \beta_1 {\bm m}_{n-1} + (1 - \beta_1) {\bm g}_n \\
{\bm v}_n & = \beta_2  {\bm v}_{n-1} +  (1 - \beta_2) {\bm g}_n^2  \\
{\bm \theta}_n & = {\bm \theta}_{n-1} - \alpha {\bm m}_n / (\sqrt{ {\bm v}_n } + \epsilon ) ,
\end{align*}
where ${\bm g}_n = {\bm \nabla} f ( {\bm \theta}_{n-1} )$, and the default values are $\beta_1 = 0.9$, $\beta_2 = 0.999$ and $\epsilon = 10^{-8}$. 
Note that all of these algorithms were introduced in the context of SGD but can be used in GD as well.

Finally, there are a number of heuristics, both with respect to learning rate schedules
\citep[e.g., "$1/t$" decay,][]{Li2020} and other modified gradients \citep[e.g., norm clipping,][]{pascanu2013}.
Keep in mind, the comparisons here are only for straight optimization (i.e., batch training), and some of these methods are perhaps best applied to mini-batch training used in SGD.
For example, the "$1/t$" decay would have $\alpha \propto 1/t$, but it will be shown on a certain example 
the best results occur when $\alpha$ increases to {\em large} values.
For these reasons comparisons will not be made to those approaches.

The methods mentioned above cover the main themes of optimization by first-order gradient-descent.
This is an active field, and there are continual innovations being introduced.

\section{The Diagnostic Metric $\rho$}

\begin{figure}[h!]
\begin{center}
\includegraphics[scale=0.32]{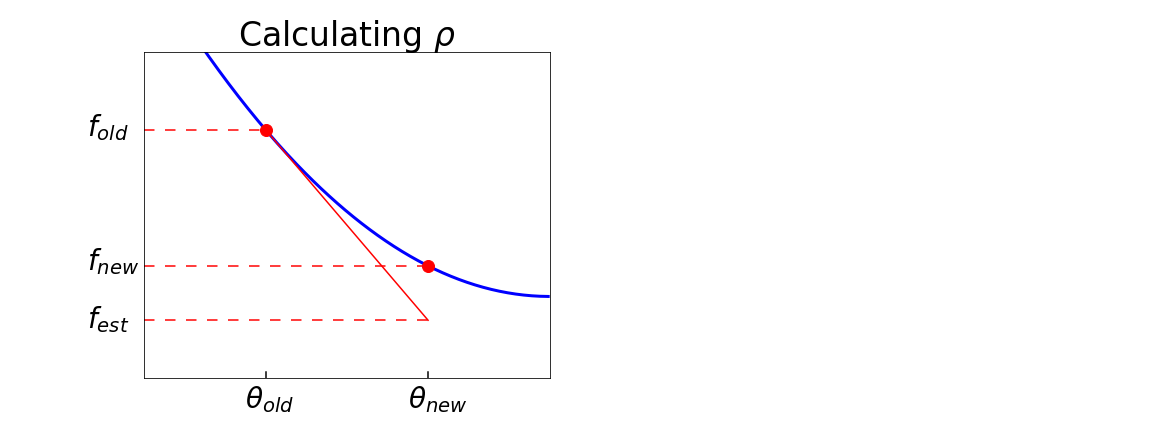}
\captionof{figure}{\color{Green} A depiction of the variables needed to compute $\rho$ when ${\bm \theta}$ is one-dimensional.}
\label{fig:rho}
\end{center}
\end{figure}

The metric introduced in this section is motivated by a focus on the accuracy of the linear approximation to the underlying function $f$.
Note that although GD is normally described as an updating of ${\bm \theta}$, it's equivalent to a linear approximation of $f$.
The generic picture of the GD update on a function is given in Fig.~\ref{fig:rho}, 
in which the parameter is updated from $\theta_{old}$ to $\theta_{new}$.  
The estimated CF according to GD is thus written
\begin{equation*}
f_{est}  = f_{old} +  {\bm \nabla} f^T d{\bm \theta} \, ,
\end{equation*}
where $f_{old} = f( {\bm \theta}_{old})$, $f_{new} = f( {\bm \theta}_{new})$, and $d{\bm \theta} = {\bm \theta}_{new} - {\bm \theta}_{old}$.
Note that with $f_{est}$ written as a function of $d\theta$, it represents a linear approximation to $f$ at $\theta_{new}$.
At the value $\theta_{new}$, the deviation from the true value is $| f_{new} - f_{est} |$.
However, to use this as a measure of the deviation from $f_{est}$, it should be made dimensionless.
The most convenient measure for this is $ | f_{old} - f_{est} |$.
Together, they are used to define the following diagnostic metric $\rho$
\begin{equation}
\rho = \left| \frac{  f_{new} - f_{est}  }{ f_{old} - f_{est} } \right| .
\label{eqn:rho}
\end{equation}
By construction, $\rho$ provides a relative measure of the error in approximating $f$ using $f_{est}$.
Importantly, for fixed $d\theta$, it is both scale and translation invariant
\footnote{An explicit dependence upon ${\bm \nabla} f$ is suppressed.}
with respect to $f$:
\begin{align*}
\rho(d\theta; f) & = \rho(d\theta; cf) \\
\rho(d\theta; f) & = \rho(d\theta; f + c) .
\end{align*}
These properties are important so that $\rho$ can be used as a universal measure across different applications.
For example, the interpretation of $\rho = 0.2$ in one application will have the same meaning as in another application.

If one does a Taylor series expansion, the leading terms comprising $\rho$ for small $d{\bm \theta}$ yield
\begin{equation*}
\rho \rightarrow \frac{1}{2} \left| \frac{ d{\bm \theta}^T ( {\bm \nabla} {\bm \nabla} f ) d{\bm \theta} }{ {\bm \nabla} f^T d{\bm \theta} }  \right| \sim {\cal O}(\alpha) \, .
\end{equation*}
These results apply when $f_{est}$ is computed using the usual first-order GD estimate.
This relationship will become useful when control algorithms are developed later in the paper.
The Hessian in the numerator also reveals that $\rho$ implicitly takes into account the local curvature of $f$, if it wasn't already obvious from the above figure.  
(However, $\rho$ is not a direct measure of the curvature.)
Also, note that there is nothing intrinsically limiting the use of $\rho$ to a {\em first-order} GD algorithm.  
It can be used within the context of second-order algorithms as well.

The metric used in trust regions (cf. Eq.~\ref{eqn:trustmetric}) is similar in idea but different in implementation compared to $\rho$ in Eq.~\ref{eqn:rho}.
The primary difference is that the trust model is second-order, whereas GD is first order.
However, ignoring that difference and taking
${\bm m}( {\bm 0} ) = f( {\bm \theta} ) = f_{old}$, ${\bm m} ( {\bm p} ) = f_{est}$, and $f( {\bm \theta} + {\bm p} ) = f_{new}$, it follows
\begin{align*}
\rho & = | 1 - r_n | .
\end{align*}
Thus the stated target range for $r_n$ of $0.25-0.5$ is seen to be quite a bit more aggressive than the target for $\rho$ used here, which is $0.1$.

\section{Algorithms}

It will be shown in this section how to compute the learning rate $\alpha$ as a function of the metric $\rho$.
This can be done exactly for simple cases, and approximately for complex ones.
Adjusting $\alpha$ to keep $\rho$ at a suitably small target value ($\rho_{targ}$) is one of the main themes of this paper.
This approach will be referred to as the  {\em constant $\rho$ ansatz}.

When it's possible to maintain $\rho$ exactly equal to $\rho_{targ}$, it's called the \underline{Ideal} case.
When it's only possible to maintain $\rho$ in an interval $(\rho_{min}, \rho_{max})$, which includes the value $\rho_{targ}$, 
it's referred to as \underline{Near-Ideal}.
The Neograd family of algorithms, which are introduced below, rely on an approximate determination of
$\alpha(\rho)$, and are Near-Ideal.  

It is important to understand the reasons for keeping $\rho$ at a small but non-zero value.
If $\rho$ is {\em too small}, the linear approximation becomes excellent, but the progress in decreasing $f$ becomes very slow.
However, if $\rho$ is {\em too large}, the update progress may improve, but it may also become erratic.
Through experience, the author has found good results with the values $\rho_{targ} = 0.1$ and $(\rho_{min},\rho_{max}) = (0.015.15)$.

Note how the philosophy of this approach is opposed to that of line-search, which attempts to decrease $f$ as much as possible in a single update.
Instead, the strategy here is that of mandating the linear approximation (i.e., GD) not make errors in approximating $f$ that are too large.
In this sense, Neograd employs a type of  trust region strategy.

\subsection{Ideal Learning Rate}

When attempting to build intuition with anything new, it is always a good idea to begin with simple examples.
For this reason, $\rho$ is computed for the Quadratic, Quartic, and Ellipse cost functions.
For the Quadratic and Quartic cases, the parameter space is n-dimensional: ${\bm \theta} = (\theta_1, ..., \theta_n)$;
for the Ellipse it is only 2-dimensional.
Using the definition of $\rho$ in Eq.~\ref{eqn:rho}, the results in Table~\ref{tab:exact} can be easily found.
\begin{table}[h!]
\centering
\vspace{5mm}
\renewcommand{\arraystretch}{1.2}
\begin{tabular}{| l | c | c | } 
\cline{2-3} 
\multicolumn{1}{c|}{} & $f$ & $\alpha$  \\
\hline
Quadratic & $c{\bm \theta}^2$ & $\rho/c$ \\
\hline
Quartic & $c{\bm \theta}^4$ & $\rho / (6c{\bm \theta}^2)$ \\
\hline
Ellipse & $Q_2$ & $\rho ( Q_4 / Q_6 )$ \\
\hline
\end{tabular} 
\caption{$\alpha$ values which lead to a constant value of $\rho$, shown for three CFs.}
\label{tab:exact} 
\end{table}
Also, ${\bm \theta}^2 = {\bm \theta} \cdot {\bm \theta}$ and the variable $Q_m$ is defined as
\begin{align*}
Q_m & = \frac{\theta_1^2}{a^m} + \frac{\theta_2^2}{b^m} \, .
\end{align*}
Although simple, these examples permit something which hasn't been done before with GD: determining the learning rate
as the function of a metric in an exact
\footnote{Only the approximate value for $\alpha$ for the Quartic case is shown; its full solution involves a solution of a cubic and doesn't really add value for our purposes here.}, 
a priori, manner.
In addition, there is a lesson in the Quartic case, in that in order to maintain $\rho$ at a fixed value, $\alpha$ must be 
increased to large values as $\theta$ goes to zero, which is the minimum of that CF.  In the example of digit recognition,
a similar phenomenon will be shown as well.

\begin{figure}[h!]
\begin{center}
\includegraphics[scale=0.32]{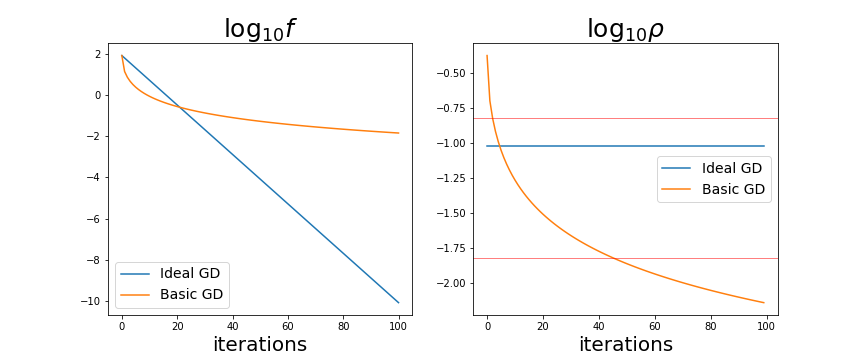}
\captionof{figure}{\color{Green} A comparison of Ideal GD using the learning rate from Table~\ref{tab:exact} to that of the usual GD, here called Basic GD.  Note the plateau in the CF value in the left plot, which is accompanied by a drop in $\rho$, as shown in the right plot.
The horizontal red lines in the right plot represent $\log \rho_{min}$ and $\log \rho_{max}$.}
\label{fig:quartic-comp-ideal}
\end{center}
\end{figure}

When one of these expressions is used in a generic GD algorithm, it is referred to as "Ideal GD".
To illustrate the improvements due to the $\alpha$ formula from Table~\ref{tab:exact}, it is compared to basic GD on the Quartic CF in Fig.~\ref{fig:quartic-comp-ideal}
The left figure shows a dramatic improvement in lowering the CF value when using this exact value.
Basic GD shows a "plateau", in which it shows little progress for higher iterations.
The right figure displays the values of $\log \rho$ for  both algorithms, and reveals that $\rho$
dramatically decreases for Basic GD.
Both of these features, the plateau in $\log f$ values and the drop of $\rho$, occur commonly when Adam is used,
and will be seen in the examples to follow.  From this, the plateau is understood to have resulted from
low $\rho$ values, which is associated with inefficient updates.

\subsection{Near-Ideal Learning Rate}

In any real application, it will not be possible to exactly solve for $\alpha$ as a function of $\rho$, as was done in the previous section.
Instead, the approach here will be to obtain an approximate expression relating $\alpha$ and $\rho$.
First, the leading order $\alpha$-dependence of the numerator and denominator of $\rho$ in Eq.~\ref{eqn:rho} is factored out:
\begin{align*}
f_{new} - f_{est}  & = A \alpha^2\\
f_{est} - f_{old}  & = B \alpha \, .
\end{align*}
With these definitions, the equation for $\rho$ becomes
\begin{equation}
\rho = \left| \frac{ \alpha A }{ B } \right| .
\label{eqn:rhoAB}
\end{equation}
In a GD algorithm, $f_{old}$ and $f_{new}$ are already available, and $f_{est}$ is easily found.
Thus, $A$ and $B$ involve little additional computation.
Observe that to lowest order in $\alpha$, $A$ is a constant.  However it does in general retain $\alpha$-dependence.
Also, $B$ is a constant; it equals $- \| {\bm \nabla} f \|^2$ and has no $\alpha$-dependence.

\subsubsection{The Neograd Algorithms}
\label{NeogradAlgos}

The planned use of Eq.~\ref{eqn:rhoAB} is the following.
After the $n$th iteration of an optimization run, exact values will be known for the entries in Eq.~\ref{eqn:rhoAB}.
Denoting them with $n$ subscripts, they appear as
\footnote{To be clear, $\rho$ is computed as a function of $\alpha$ in each iteration.  It is $\rho$, not $\alpha$, that is the dependent variable}
\begin{equation}
\alpha_n =  \left| \frac{B_n}{A_n} \right| \rho_n \, .
\label{eqn:alphan}
\end{equation}

In an ideal world, the $\rho_n$ would already equal the target value of $\rho_{targ}$.
However, since that's likely not the case, one can instead use the above equation to find an $\alpha$ for the next iteration that should produce a $\rho$ 
closer to $\rho_{targ}$:
\footnote{It can be shown that this process will in generic circumstances tend to converge on an $\alpha$ that leads to $\rho_{targ}$.  See Appendix C.}
\begin{equation}
\alpha_{n+1} =  \left| \frac{B_n}{A_n} \right| \rho_{targ} \, .
\label{eqn:alphan1}
\end{equation}
This will be referred to as the {\em Adaptation Formula} (AF).
This scaling down of Eq.~\ref{eqn:alphan} is of course approximate, since $A$ in general retains $\alpha$-dependence.
In any event it is this value that will be used as the learning rate in the {\em next} iteration.
This is the essence of the here-defined Neograd family of algorithms.

There are two refinements that can be made to this approach.  
The first is based on the recognition that Eq.~\ref{eqn:alphan1} will certainly work better if the difference between $\rho_n$ and $\rho_{targ}$ is small
\footnote{That is, the map has a finite radius of convergence about $\rho_{targ}$.  See Appendix C}.  
Thus, as shown in Algo.~\ref{algo:get-rho-prime} in Appendix A, it may be advantageous to try to change $\rho$ in several steps 
(e.g., $\rho \rightarrow \rho' \rightarrow  \rho_{targ})$.
The second refinement is to hybridize this family of algorithms with existing techniques, such as momentum.
It is this variant, called NeogradM, which the author has found most performant.
\footnote{It is especially helpful in cases where the CF has narrow valleys in the shape of the CF}
.

Finally, there is an additional variation in the manner in which $f_{est} = {\bm \nabla} f \cdot d{\bm \theta}$ is computed due to the use of momentum.
Denoting the momentum for ${\bm \nabla} f$ by ${\bm m}$, 
there are two
\footnote{In addition, one might also consider momentum on $d{\bm \theta}$ rather ${\bm \nabla} f$.  The distinction arises because $\alpha$ and ${\bm \nabla f}$ change, and $d{\bm \theta}$ depends on both.}
obvious options for computing $f_{est}$:
\begin{enumerate}
\item $f_{est} = {\bm \nabla} f \cdot (-\alpha {\bm m})$ 
\item $f_{est} = -\alpha m^2$ 
\end{enumerate}
Option \#1 is natural for $f_{est}$ since ${\bm m}$ was used for the update of ${\bm \theta}$, so it is an exact calculation of $f_{est}$.
However, given that $d{\bm \theta}$ is based on ${\bm \nabla} f$, and that ${\bm \nabla} f$ really appears twice in $f_{est}$,
one might also consider option \#2 and use momentum on both appearances of ${\bm \nabla} f$.
The results of this paper are based on option \#1, with the exception of Sec.~\ref{sec:fest-alt} which uses option \#2.

\section{Experiments}

In this section two test functions and an application are used to evaluate Neograd.
The author advocates for the use of test functions, as they permit a window into the exact functioning of the algorithm.
Also, it is arguably the case that an application will at times be limited by characteristics of a CF that are well-described by one
or another such test functions.

Neograd is primarily compared to Adam, a popular choice for optimization in the field of ML.
An effort was made to find the best learning rate for Adam, while its other parameters were set to their default values.
In comparison, this was not necessary with Neograd, since the algorithm itself suggests an easy way to set $\alpha$ (see Appendix B).

\subsection{Demonstration of the Adaptivity of $\alpha$}
\label{sec:sigmoid-well}

Before examining the performance of the NeogradM algorithm, it is worthwhile to first understand how it adjusts the learning rate
in the face of very flat and steep CF profiles.
To that end, it is used on the {\em sigmoid-well} CF, defined by
\begin{align*}
f(\theta) & = \sigma [ s(-\theta - a) ]  + \sigma [ s(\theta - a) ] \\
\sigma(\bullet) & = 1/(1 + e^{-\bullet} ) \, ,
\end{align*}
where $s=10$ and $a=2$.
\begin{figure}[h!]
\begin{center}
\includegraphics[width=0.6\linewidth]{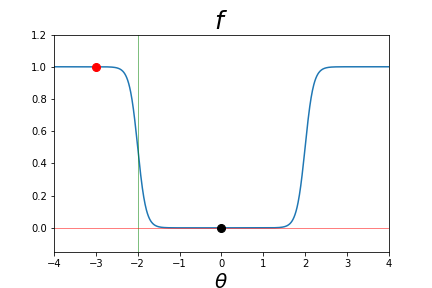}
\captionof{figure}{\color{Green}  Profile of the CF {\em sigmoid-well}.  Updates begin at $\theta=-3$ (red dot) and approach the minimum at $\theta=0$ (black dot).  The inflection point in the CF is marked by a vertical green line.}
\label{fig:expt-sigmoid-well-1}
\end{center}
\end{figure}
As show in Fig.~\ref{fig:expt-sigmoid-well-1}, the initial parameter value is $\theta = -3$, and is marked by a red dot; 
the minimum occurs at $\theta=0$ and is marked by a black dot.
Also, note the inflection point in the CF is near $\theta = -2$.

\begin{figure}[h!]
\begin{center}
\includegraphics[width=1.0\linewidth]{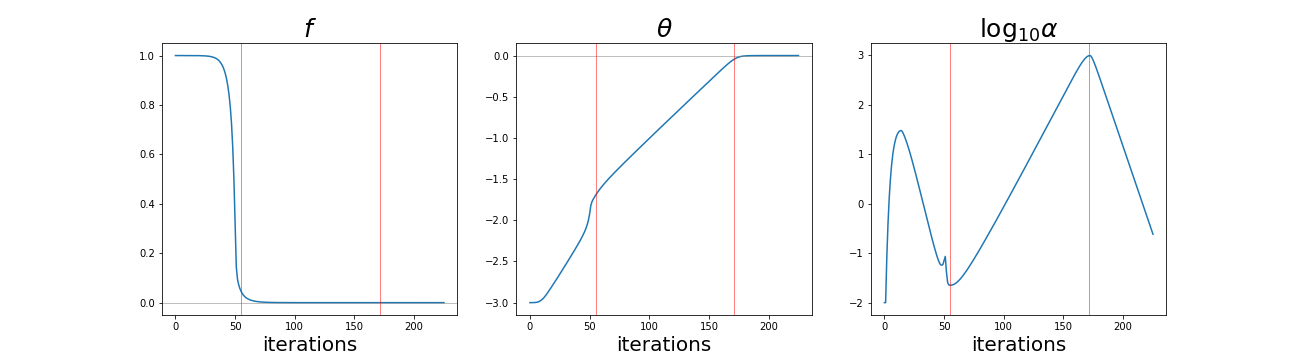}
\captionof{figure}{\color{Green} Comparative plots of the CF, $\theta$, and $\log \alpha$.  Vertical lines are added to the plots to indicate turning points in the behavior of $\alpha$, which are shown in the right plot.}
\label{fig:expt-sigmoid-well-2}
\end{center}
\end{figure}

As the updates move the current $\theta$ from the red to the black dot, there are two significant turning points,
marked by vertical red lines in Fig.~\ref{fig:expt-sigmoid-well-2}; they occur near iterations 55 and 171.
The most important changes are displayed in the right plot of Fig.~\ref{fig:expt-sigmoid-well-2},
which shows (after an initial increase) $\alpha$ decreasing up until the 55th iteration, increasing up until the 171st iteration, and decreasing thereafter.
With respect to the left plot in the same figure, this is understood by examining the local CF profile of the current $\theta$.
The initial decrease corresponds to when $\theta$ is approaching the steep drop; hence it must make smaller updates.
After it passes that point, it must then increase $\alpha$ to reach the minimum, which it does between iterations 55 and 171.
At this point it's very close to $\theta=0$, and it changes over to $\alpha$ decreasing once again.
This demonstrates how NeogradM is capable of automatically adjusting $\alpha$ to properly deal with a changing landscape.

\subsection{Beale's Function}
\label{sec:Beale}

\begin{figure}[h!]
\begin{center}
\includegraphics[width=1.1\linewidth]{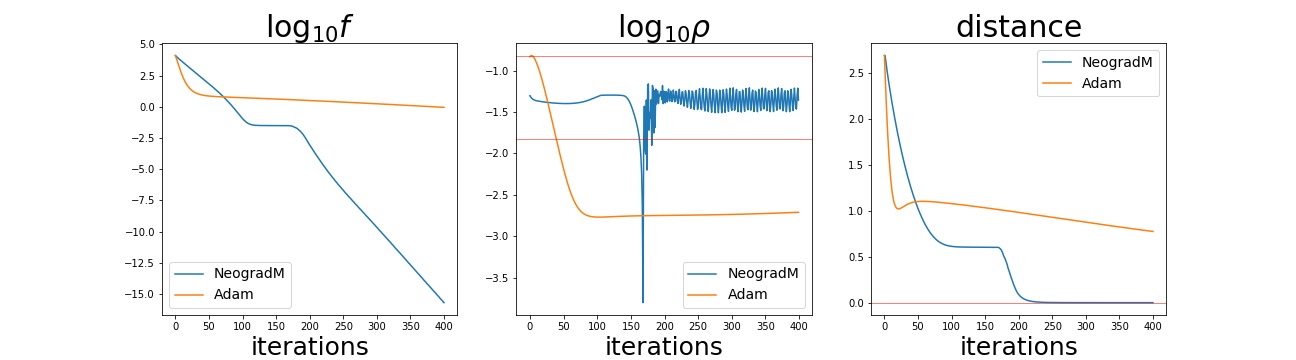}
\captionof{figure}{\color{Green} Comparison plots between NeogradM (blue) and Adam (orange).  Of note is that NeogradM reaches a low CF value much more quickly (left plot), which mirrors its relatively high $\rho$ values (center plot).  NeogradM is thus able to close the distance to the minimum much more quickly than Adam (right plot).  The horizontal red lines in the center plot correspond to $\log\rho_{min}$ and $\log\rho_{max}$.}
\label{Beale-triple}
\end{center}
\end{figure}

Beale's function is used as a standard test for optimization algorithms, as it offers several challenging features in its CF landscape.
It is defined as 
\begin{equation*}
f = (1.5 - \theta_1 + \theta_1 \theta_2 )^2 + (2.25 -\theta_1 + \theta_1 \theta_2^2)^2 + (2.625 - \theta_1 + \theta_1 \theta_2^3)^2 \, .
\end{equation*}
It has a global minimum at $(\theta_1,\theta_2)=(3,0.5)$.  The initial condition used here is $(4,3)$, from which the minimum is reachable.
In Fig.~\ref{Beale-triple} are comparison plots between Adam and NeogradM when applied to this CF.
NeogradM reaches a much lower value than Adam in the $\log f$ plot; the exact amount depends on how long the run is.
Also, the plot of $\log \rho$ shows that Adam again naturally evolves into a situation where it produces very small $\rho$ values,
which as discussed earlier is indicative of overly conservative and slow progress.
The graph of $\log \rho$ due to NeogradM does a good job of staying within the red lines (i.e., $\log\rho_{min}$ and $\log\rho_{max}$), 
except for a narrow spike to a low value.
Recall that such overly small $\rho$ values are merely inefficient, whereas overly large values are potentially problematic.
This deviation seems to be a response to a very flat region in the CF profile.
Finally, the right plot in the figure shows that NeogradM has reached its target after about 250 iterations, at which point Adam
is still far from it.

\begin{figure}[h!]
\begin{center}
\includegraphics[width=0.7\linewidth]{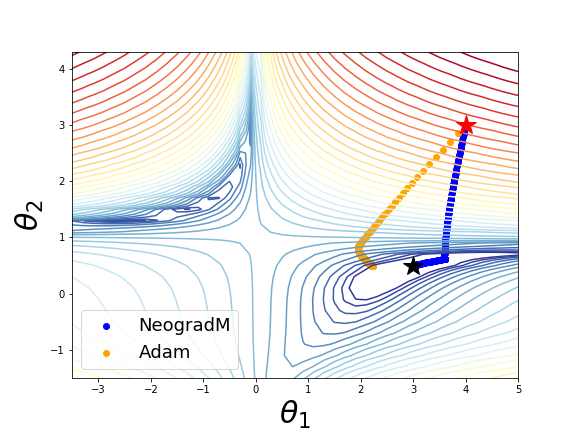}
\captionof{figure}{\color{Green} 
With a contour plot of the Beale's function as the background, the ${\bm \theta}$-paths for Adam and NeogradM are displayed.
Note how the slower path of Adam has a significant component parallel to the contour lines.
The red (black) star is the initial (final) point in this experiment.
}
\label{Beale-contour}
\end{center}
\end{figure}

Finally, it is illuminating to examine a plot of the updates of each algorithm as they move through the CF landscape.
In Fig.~\ref{Beale-contour} a red star indicates the starting point for the algorithms, and a black star the minimum.
The path taken by NeogradM (the blue dots), takes a very direct route; it is nearly perpendicular to the contour lines.
Recall that a pure GD algorithm would be perfectly perpendicular; the slight discrepancy is due to the momentum in NeogradM.
In contrast, Adam takes a relatively oblique path toward the black star; it is clearly {\em not} perpendicular to the contour lines.
As discussed in Sec. 2.1 of \citep{neograd_arxiv_v2}, Adam utilizes preferential weighting which adds significant weight to the null space
term in the solution for $d{\bm \theta}$, which leads to updates parallel to the contour lines.

\subsection{Cross Entropy Penalty (digit recognition)}

The purpose of this section is mainly to demonstrate the performance of NeogradM in lowering the training error.  
The reader should not confuse this goal with that of lowering the generalization error, which the author considers to be qualitatively different since it also involves regularization. 
Later in this section, additional comparisons will be made between NeogradM and Adam, in order to offer insight into the different workings of the two algorithms.

\begin{figure}[h!]
\begin{center}
\includegraphics[width=0.8\linewidth]{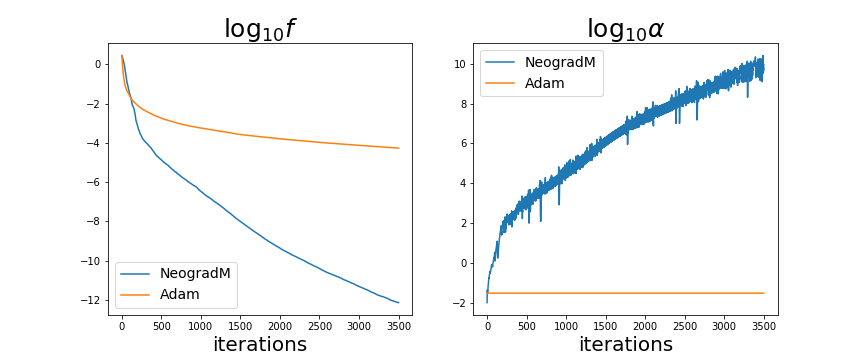}
\captionof{figure}{\color{Green} 
In the left plot, note the 8 order of magnitude improvement by NeogradM over Adam for $\log f$.
In the right plot, observe how $\alpha$ increases to very large values as the optimization proceeds.
(The value of $\alpha$ used in Adam is included for reference.)}
\label{fig:digits-triple}
\end{center}
\end{figure}

The training error of interest is based on the cross entropy penalty of a fully connected, single-hidden layer NN meant for classifying digitized images of the numbers 0 through 9 \citep{scikit-digits,Dua2019}.  For this model, the dimensionality of ${\bm \theta}$ is 2260.
\footnote{The NN consisted of 64 input nodes, 30 hidden nodes, and 10 output nodes.  Also, 1437 labeled training images were used.  A tanh activation function was used on the hidden layer, and a softmax on the final layer.  The model had no regularization.}
Both Adam and NeogradM were used in minimizing this CF using 3500 iterations, resulting in Fig.~\ref{fig:digits-triple}.
The most noticeable feature in the $\log f$ plot is that NeogradM leads to a significantly smaller value of the CF, by a factor of $10^8$.
This result is taken as evidence of the advantages NeogradM has over existing algorithms, such as Adam. 
Associated with that is the plot of $\log\alpha$, which reveals the dramatic increase of $\alpha$ throughout the course of the optimization.
(In this figure the value of $\log\alpha$ for Adam is shown for reference
\footnote{It would perhaps be a fairer comparison to determine some sort of average measure of the different effective learning rates for each component of ${\bm \theta}$ in the Adam algorithm.  However, that's beyond the scope of this section.} 
.)
This plot shows that $\alpha$ needs to be increased in order to keep $\rho$ in the target interval.

\begin{figure}[h!]
\begin{center}
\includegraphics[width=0.8\linewidth]{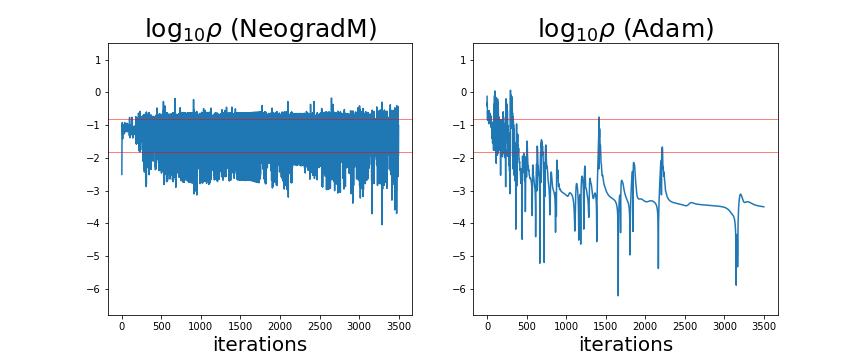}
\captionof{figure}{\color{Green} Illustration of $\log_{10}\rho$ for NeogradM (left) and Adam (right).  
The red lines correspond to $\log\rho_{min}$ and $\log\rho_{max}$.}
\label{fig:digits-rho}
\end{center}
\end{figure}

Next, in Fig.~\ref{fig:digits-rho}, a side-by-side comparison is made of $\log\rho$ for the two algorithms.
As expected, NeogradM does a fairly good job of keeping $\rho$ in the target interval,
whereas Adam falls far below it.  This is similar to what was seen before with Adam.
It also reveals that in this application it's necessary to adjust $\alpha$ rather frequently,
as is clear from the raggedness of the plot of $\log \rho$ for NeogradM.
These plots of $\log \rho$ are shown since this metric plays an important role in NeogradM, and also because the low-$\rho$ values of Adam are being identified as the reason for that algorithm having a plateau in the CF vs. iteration plots.

\subsubsection{Basins of Attraction}

\begin{figure}[h!]
\begin{center}
\includegraphics[width=1.0\linewidth]{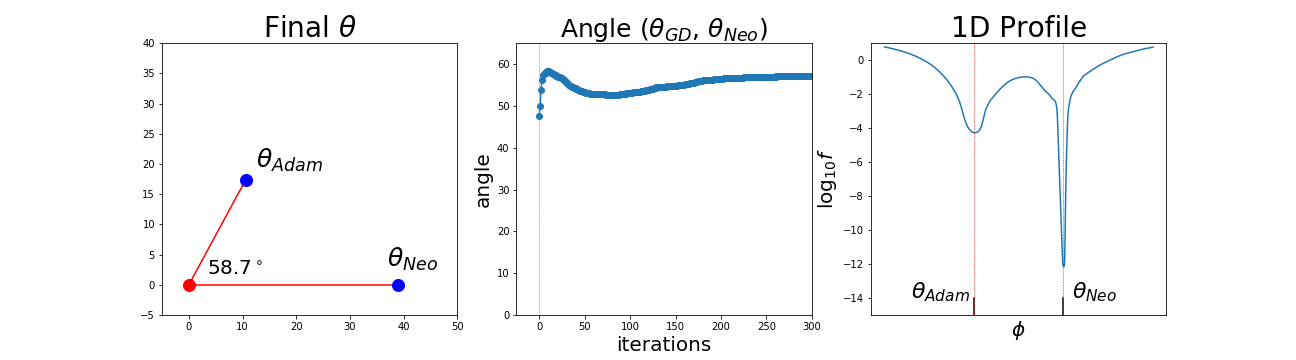}
\captionof{figure}{\color{Green} 
The left plot illustrates the differences in the magnitude and direction for the final parameter values for Adam and NeogradM
(i.e., ${\bm \theta}_{Adam}$ and ${\bm \theta}_{Neo}$); this indicates the algorithms found very different minima.
The center plot shows this angular difference vs. iterations, and reveals that it largely appears after a single iteration.
The right plot shows the profile of $\log f$ on a line connecting the two final ${\bm \theta}$ values in ${\bm \theta}$-space; it shows they are well-separated basins of attraction, at least along this line.}
\label{fig:thetas-comparison}
\end{center}
\end{figure}

To further investigate the differences in the paths from the ${\bm \theta}$-updates of these two algorithms,
three plots in Fig.~\ref{fig:thetas-comparison} are presented.
The left plot is a 2D comparison of the two final parameter vectors, ${\bm \theta}_{Adam}$ for Adam and ${\bm \theta}_{Neo}$ for NeogradM.
The angle between them is larger than might be expected, equalling over 58 degrees.
To understand how this difference might result, a plot of the angle between ${\bm \theta}_{Adam}$ and ${\bm \theta}_{Neo}$
is given as a function of iterations.
Somewhat surprisingly, almost all of the the change in direction happens after the first iteration.
This reflects how Adam, by construction, does not follow the direction of steepest descent \citep[cf.][]{Wilson2017}.
This was seen previously in the section on Beale's function \ref{sec:Beale}, where the updates had a significant component parallel to the contour lines of the CF.
Finally, the third plot illustrates the differences in basins of attraction reached by each algorithm.
(Keep in mind that this comparison is along a single dimension; the underlying dimensionality of the space is 2260.)
The interpolating vector between the two ${\bm \theta}$s is defined as 
\begin{equation*}
\phi = (1-s) {\bm \theta}_{Adam} + s{\bm \theta}_{Neo} \, .
\end{equation*}
In this plot the range of $s$ is $[-1,2]$.  
This plot shows the basins are distinct (at least along this one dimension).
While it may seem that Adam has simply discovered a higher, slightly wider basin, it's actually the case
that it simply hasn't followed the gradient of $f$ down to lower values.
To wit, if Adam were allowed to continue 
\footnote{For this "continuation experiment", momentum variables were re-initialized to zero at the start of the run, for both Adam and NeogradM.  The additional experiment of running Adam for 7000 iterations resulted in $\log f \approx -5.08$.  Thus the restart slightly benefited Adam.} 
the optimization for an additional 3500 iterations, it would only have gone from $\log_{10} f = -4.27$ to $-5.47$.
However, if NeogradM had instead continued from where Adam left off at 3500 iterations, it would have gone from $\log_{10} f = -4.27$ to $-9.28$.
This shows it's not that Adam somehow found a shallower basin, it's that it wasn't as effective in reaching a lower value in that basin.

\subsubsection{Speedup}

\begin{figure}[h!]
\begin{center}
\includegraphics[width=1.0\linewidth]{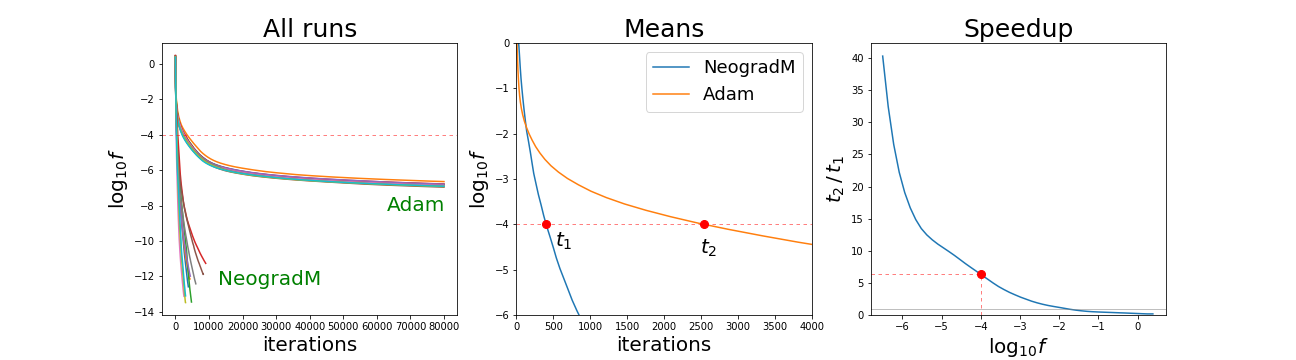}
\captionof{figure}{\color{Green} 
The left plot illustrates the 10 runs each for NeogradM and Adam; note the plateau in the upper grouping for Adam.
The center plot uses the means from the 10 runs, and also illustrates how $t_1$ and $t_2$ are defined 
at the example value of $\log f = -4$ (cf. red dashed line).
In the right plot, the speedup is plotted vs. $\log f$.  Note its dramatic increase when Adam reaches its plateau.}
\label{fig:digits-speedup}
\end{center}
\end{figure}

In this section the speedup achieved by NeogradM over Adam is quantified.
The speedup is defined by the ratio $t_2/t_1$,
where $t_1$ [$t_2$] is the number of iterations needed by Adam [NeogradM] to reach a specified CF value.
\footnote{Another way to measure the performance increase is to compare CF values for a given iteration.}
To reduce variation in the results due to the different initial ${\bm \theta}$s, 10 runs were made.
In Fig.~\ref{fig:digits-speedup}, three plots are shown of the results of this experiment.
Adam was run for 80000 iterations, while NeogradM was run for much shorter times (until the limits of machine precision caused it to stop).
In the left plot, all runs are plotted  
\footnote{Although it seems like NeogradM has a higher variance (along the iteration axis),
Adam actually has a higher variance for a given CF value because of its strong plateau.}
.
The center plot was obtained by computing the average number of iterations for a fixed CF value for each algorithm.
It also gives a visual depiction of how the speedup is defined in this close-up view.
Finally, in the right plot the speedup is computed for a larger range of the CF, and it reveals that
in reaching these lower CF values, NeogradM is {\em much} faster than Adam.
It is a reflection of the effective plateau in CF values produced by Adam, which seems to occur near $f \approx 10^{-7}$.
Near the plateau, the speedup appears to become arbitrarily large, becoming $\approx 40$ in the figure.

\subsubsection{Alternate Formulation of $f_{est}$}
\label{sec:fest-alt}

\begin{figure}[h!]
\begin{center}
\includegraphics[width=0.55\linewidth]{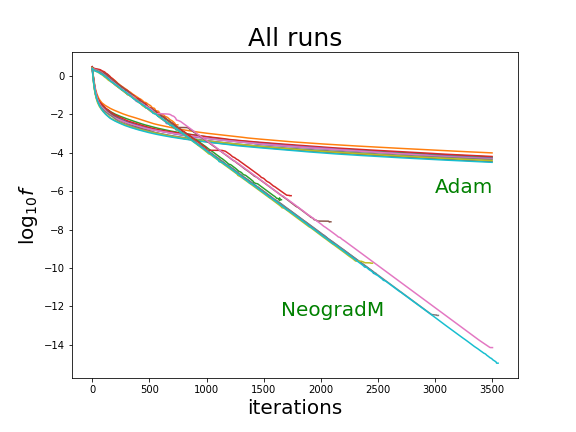}
\captionof{figure}{\color{Green} 
10 runs were conducted for both NeogradM and Adam, using the alternate formulation of $f_{est}$ (cf. Eq.~\ref{eqn:fest-alt}).
Perhaps the most striking difference is that the graph for NeogradM assumes an approximate linear form.
}
\label{fig:fest-alt}
\end{center}
\end{figure}

At the end of Sec.~\ref{NeogradAlgos} an alternate formulation (i.e., option \#2) of the estimated CF value $f_{est}$ was discussed, one in which it would be computed as
\begin{align}
f_{est} & = {\bm \nabla} f \cdot d{\bm \theta} = -\alpha m^2
\label{eqn:fest-alt}
\end{align}
where ${\bm m}$ is the momentum variable for ${\bm \nabla} f$, and $m^2 = {\bm m} \cdot {\bm m}$.
Using this formulation, and setting the momentum parameter at $\beta = 0.9$, the result for the $\log f$ versus iterations was computed as shown in Fig.~\ref{fig:fest-alt},
alongside the same results from Adam.
This numerical experiment was conducted for 10 different initial conditions for each algorithm.
Each run was continued until the limits of machine precision forced the computation to cease.
Note that the lengths of each run varied over a wider range than for the previous runs with NeogradM (cf. Fig.~\ref{fig:digits-speedup}) or for Adam.
Aside from the differences in performance, perhaps the most noticeable feature of the graph for NeogradM is that $\log f$ varies approximately {\em linearly} as a function of the number of iterations.
This is in sharp contrast to the graph from Adam, as well as to the graph for NeogradM in Fig.~\ref{fig:digits-speedup} (the left plot), 
where $f_{est}$ was computed using option \#1.
The reader should note that such behavior (partially) appeared in a similar plot (Fig.~\ref{Beale-triple}) for the Beale's CF (which also used option \#1).
Also, in a previous draft of this paper \citep{neograd_arxiv_v2}, such linear behavior was derived for two simple CFs (quadratic and quartic)
using the prescription for $\alpha$ listed in Table~\ref{tab:exact}.
In general though, the author obtained better results from option \#1 than option \#2.
The task of showing why such a linear dependence appears for the choice of $f_{est}$ in Eq.~\ref{eqn:fest-alt} 
on this relatively complex example is outside the scope of the present paper.

\section{Final Remarks}

The purpose of this paper was defined by two goals.
The first was to devise an algorithm that didn't suffer from a plateau in the graph of the CF vs iterations.
The algorithm NeogradM achieves just that: in Fig.~\ref{Beale-triple} it had a lower CF value by a factor of $10^{15}$ compared to Adam on Beale's function.
On the cross-entropy function in Fig.~\ref{fig:digits-triple}, it had a lower CF by a factor of $10^8$.
The second goal was to efficiently determine a good value for $\alpha$.
Here, the value for $\alpha$ was determined as an approximate function of $\rho$.
Adjusting $\alpha$ to keep $\rho$ at a fixed small value (i.e., the {\em constant $\rho$ ansatz}) was done done through the  
Adaptation Formula (Eq.~\ref{eqn:alphan1}) and formed the basis of the Neograd family of algorithms.
These ideas were successfully demonstrated on the examples in the {\em Experiments} section.

In the section {\em Related Work}, a metric for Trust Region methods was recalled.
It differed from the metric proposed here ($\rho$) in that it was based on a quadratic, rather than a linear model.
The idea for Neograd was to use the linear extrapolation of GD only when deviations from it were relatively small; this corresponds to a small $\rho$.
In this sense it is similar to Trust Region methods, which sought to keep a metric $r_n$ small by adjusting the trust radius $\Delta_n$ \citep[cf. Algorithm 4.1 in][]{Nocedal-book2006}.  However, rather than using something like the Adaptation Formula (Eq.~\ref{eqn:alphan1}) as a guide, the Trust Region approach resorts to increasing or decreasing $\Delta_n$ by factors of 2.

The approach with Neograd of trying to keep deviations from the linear model (i.e., GD) small also differs from the Line Search methods, which aim to choose an $\alpha$ which leads to the lowest possible reduction in the CF, regardless of how poor a fit the linear extrapolation may become.  
Also, the Wolfe conditions (Eq.~\ref{eqn:Wolfe}, which seek to limit $\alpha$ for different reasons, involve terms that are second-order in $({\nabla f})^2$, and hence are inequivalent to $\rho$.

\section*{Acknowledgement}

The author thanks Sebastian J.~Wetzel for his suggestions on improving the manuscript.

\section*{Appendix A: Pseudocode for Neograd Algorithms}

Pseudocode that implements the Adaptation Formula (AF) (Eq.~\ref{eqn:alphan1}) as well as a gradual change
\footnote{When this gradual change of $\rho$ is not implemented, the version is just called Neograd\_v0.}
of $\rho$ toward $\rho_{targ}$ is called Neograd\_v1 and is shown in Algo.~\ref{algo:neograd-v1}.
\begin{algorithm}[H]
\caption{: Neograd\_v1}
\begin{algorithmic}
\STATE{Input: $({\bm \theta}_{old}, \alpha, \rho_{targ}, \text{num})$ }
\STATE{ $f_{old} = f({\bm \theta}_{old})$ }
\FOR{i = 1 to num}
\STATE{ ${\bm g} = {\bm \nabla} f({\bm \theta}_{old})$ }
\STATE{ $d{\bm \theta} = -\alpha {\bm g}$ }
\STATE{ ${\bm \theta}_{new} = {\bm \theta}_{old} + d{\bm \theta}$ }
\STATE{ $f_{new} = f({\bm \theta}_{new})$ }
\STATE{ $f_{est} = f_{old} + {\bm g} \cdot d{\bm \theta}$ }
\STATE{ $\rho = \text{get\_rho}(f_{old}, f_{new}, f_{est})$ }
\STATE{ $\rho' = \text{get\_rho\_prime}( \rho, \rho_{targ} )$ }
\STATE{ $\alpha = \text{get\_alpha}(f_{old}, f_{new}, f_{est}, \alpha, \rho')$ }
\STATE{ $f_{old} = f_{new}$ }
\STATE{ ${\bm \theta}_{old} = {\bm \theta}_{new}$ }
\ENDFOR
\STATE{Return: ${\bm \theta}_{new}$ }
\end{algorithmic}
\label{algo:neograd-v1}
\end{algorithm}
In this pseudocode, {\em get\_rho} implements Eq.~\ref{eqn:rho} and {\em get\_alpha} implements Eq.~\ref{eqn:alphan1}.
This algorithm implements the idea of an intermediate change of $\rho'$, 
which is shown here as {\em get\_rho\_prime} and is implemented as Algo.~\ref{algo:get-rho-prime} below.
(This was discussed in Section \ref{NeogradAlgos}).
The motivation for this is that the AF is only expected to be stable for small deviations from $\alpha_n$ and $\rho_n$ (cf. Eq.~\ref{eqn:alphan}).
Hence, if $\rho_n$ differs too greatly from $\rho_{targ}$, trying to use $\rho_{targ}$ in the AF might not lead to the desired behavior (which is to produce an $\alpha$ which more closely leads to $\rho_{targ}$).
The stability of the AF under such iterations is discussed in Appendix C.
\begin{algorithm}[H]
\caption{: get\_rho\_prime}
\begin{algorithmic}
\STATE{Input: $(\rho, \rho_{targ})$ }
\IF{$\rho < \rho_{targ}$ }
\STATE{ $r = 0.75 \log_{10}(\rho/\rho_{targ})$ }
\STATE{ $\rho' = 10^r  \rho_{targ}$ }
\ELSE
\STATE{ $\rho' = \rho_{targ}$ }
\ENDIF
\STATE{Return: $\rho'$ }
\end{algorithmic}
\label{algo:get-rho-prime}
\end{algorithm}
For example, when $\rho = 10^{-9}$ and $\rho_{targ}$ = 0.1, {\em get\_rho\_prime} returns $\rho' = 10^{-7}$  
So with respect to log values, $\rho'$ is 3/4 of the way to $\rho$ and 1/4 to $\rho_{targ}$.
Note that when $\rho > \rho_{targ}$, no intermediate value is used, since large $\rho$ values are indicative
of uncontrolled behavior.  Thus $\rho$ should be reduced as quickly as possible, even if it means
having the AF become inapplicable.

Another improvement that can be made is to hybridize the algorithm with existing ones,
such as Adam or RMS Prop.
The ability to hybridize with such algorithms is compactly achieved in Algo.~\ref{algo:neo-hybrids} by accessing a function {\em get\_combo}
\footnote{See \citep{MZGithub} for an implementation.}
and passing a parameter {\em type-opt}  which indicates which algorithm is to be used.  
\begin{algorithm}[H]
\caption{: Hybridized Neograd}
\begin{algorithmic}
\STATE{Input: (${\bm \theta}_{old}$, $\alpha$, $\rho_{targ}$, num, type\_opt)}
\STATE{ $f_{old} = f({\bm \theta}_{old})$ }
\STATE{ ${\bm v}={\bm 0}, {\bm v2}={\bm 0}$ }
\FOR{i = 1 to num}
\STATE{ ${\bm g} = {\bm \nabla} f({\bm \theta}_{old})$ }
\STATE{ $d{\bm \theta}, {\bm v}, {\bm v2} = \text{get\_dp\_combo} ( \text{type\_opt}, {\bm g}, {\bm \theta}_{old}, {\bm v}, {\bm v2}, i, \alpha)$ }
\STATE{ ${\bm \theta}_{new} = {\bm \theta}_{old} + d{\bm \theta}$ }
\STATE{ $f_{new} = f({\bm \theta}_{new})$ }
\STATE{ $f_{est} = f_{old} + {\bm g} \cdot d{\bm \theta}$ }
\STATE{ $\rho = \text{get\_rho}(f_{old}, f_{new}, f_{est})$ }
\STATE{ $\rho' = \text{get\_rho\_prime}( \rho, \rho_{targ} )$ }
\STATE{ $\alpha = \text{get\_alpha}(f_{old}, f_{new}, f_{est}, \alpha, \rho')$ }
\STATE{ $f_{old} = f_{new}$ }
\STATE{ ${\bm \theta}_{old} = {\bm \theta}_{new}$ }
\ENDFOR
\STATE{Return: ${\bm \theta}_{new}$ }
\end{algorithmic}
\label{algo:neo-hybrids}
\end{algorithm}
In the pseudocode, the parameters ${\bm v}$ and ${\bm v2}$ are passed to this function, which are momentum variables for the gradient and gradient-squared.
This allows convenient access to GD with momentum, Adam, RMS Prop, and others.
As mentioned previously, the variant that has shown the best performance is NeogradM, which utilizes momentum on the gradient updates (i.e., ${\bm v}$).
Finally, it is also convenient to have a separate method to determine a good initial value for $\alpha$.
An example implementation is given next, in Appendix B.

\section*{Appendix B: Initial Learning Rate}

When running GD it's necessary to determine an initial learning rate $\alpha$, which normally involves doing computationally costly runs.
A much more expedient way is to use the Adaptation Formula (AF) to check a small number of trial values, computing $\rho$ for each.
The first one that leads to a value of $\rho$ within $(\rho_{min}, \rho_{max})$ is taken as the initial value.
Pseudocode that implements this idea is given in Algo.~\ref{algo:get-starting-alpha}.
\begin{algorithm}[H]
\caption{: get\_starting\_alpha}
\begin{algorithmic}
\STATE{Input: $({\bm \theta}_{old}, \alpha, \rho_{targ}, \text{nrep} )$ }
\STATE{ $f_{old} = f({\bm \theta}_{old})$ }
\STATE{ ${\bm g} = {\bm \nabla} f({\bm \theta}_{old})$ }
\FOR{j = 1 to nrep}
\STATE{ $d{\bm \theta} = -\alpha {\bm g}$ }
\STATE{ ${\bm \theta}_{new} = {\bm \theta}_{old} + d{\bm \theta}$ }
\STATE{ $f_{new} = f({\bm \theta}_{new})$ }
\STATE{ $f_{est} = f_{old} + {\bm g} \cdot d{\bm \theta}$ }
\STATE{ $\rho = \text{get\_rho}(f_{old}, f_{new}, f_{est})$ }
\IF{ $\rho \in (\rho_{min}, \rho_{max})$ } 
\STATE{Return: $ (\alpha, \rho)$ }
\ELSE
\STATE{ $\rho' = \text{get\_rho\_prime}( \rho, \rho_{targ} )$ }
\STATE{ $\alpha = \text{get\_alpha}(f_{old}, f_{new}, f_{est}, \alpha, \rho')$ }
\ENDIF
\ENDFOR
\STATE{Return: $ (\alpha, \rho)$ }
\end{algorithmic}
\label{algo:get-starting-alpha}
\end{algorithm}
The input parameter $nrep$ denotes the number of attempts used to find a suitable initial $\alpha$.
The input $\alpha$ is the starting point in this search; it is recommended to choose one overly small,
since as shown in Fig.~\ref{fig:stability-204}, nonlinear structure in $\rho (\alpha)$ can lead to relatively large $\alpha$ values that have a $\rho$ near the target $\rho_{targ}$.
Finally, the reader should note that there are a number of other ways to create such an algorithm based on the AF; this is only one example.

\section*{Appendix C: Stability of the Adaptation Formula}

In Section \ref{NeogradAlgos} the Adaptation Formula (AF) was introduced in Eq.~\ref{eqn:alphan1}, but its stability was not discussed.
Recall that the Neograd algorithm is defined by Eqs.~\ref{eqn:alphan} and \ref{eqn:alphan1}, in which the update of $\alpha$
is used to generate a new ${\bm \theta}$.
However, when those equations are used \underline{without updating ${\bm \theta}$}, they should lead to $\rho$ converging towards $\rho_{targ}$, i.e.,
\begin{align*}
d_n & = | \rho_n - \rho_{targ} | \\
d_n  & > d_{n+1} ,
\end{align*}
where $n = 1, 2, 3, ...$  
This will be demonstrated next on the sigmoid-well CF used earlier in Section \ref{sec:sigmoid-well}.

\begin{figure}[h!]
\begin{center}
\includegraphics[scale=0.32]{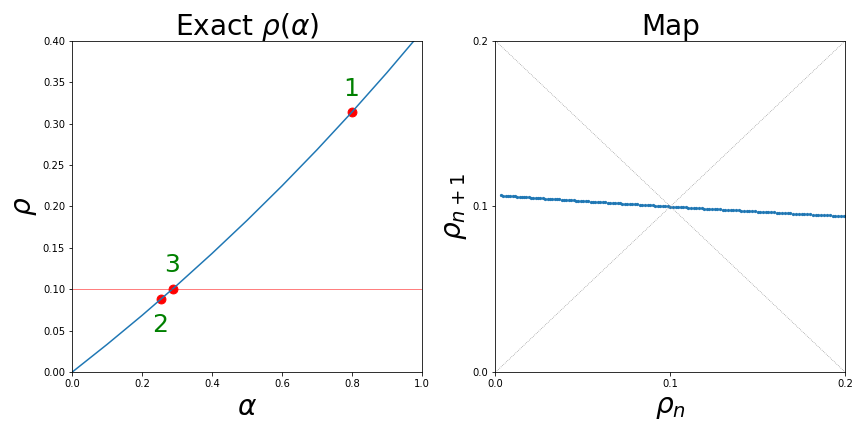}
\captionof{figure}{\color{Green} Here, with $\theta_0 = -2.5$, the iterates lead to convergence towards $\rho_{targ}$, indicated by the red line.  
In the left plot this is shown as a sequence of numbered points which are updates using the AF.
This is reflected in the $\rho$-map in the right plot, where a slope of magnitude less than 1 indicates stable updates; this plot was produced using $\alpha \in (0.01,0.8)$.
The gray lines at $\pm 45^\circ$ are added as a visual guide.}
\label{fig:stability-250}
\end{center}
\end{figure}

Using the starting point of $\theta_0 = -2.5$, $\rho$ is computed as a function of $\alpha$, as shown in the left plot of Fig.~\ref{fig:stability-250}.
The numbered points on this plot indicate the updates due to the repeated application of the AF.
Thus, one might begin at point \#1 with $(\alpha,\rho) = (0.8, 0.314)$, apply the AF to get point \#2 at $(0.255,0.088)$, and then apply the AF again
to get point \#3 at $(0.288,0.101)$.  
Once again, this is done {\em without updating ${\bm \theta}$}.
Recall that the goal of the AF is to obtain a $\rho$ close to $\rho_{targ} = 0.1$, which is shown as a horizontal red line in the plot.
Hence, in this case the AF clearly leads to convergence towards $\rho_{targ}$.
Another way to understand this progression of $\rho$ values, from one update to the next, is to display them as a map
\footnote{This kind of plot may be reminiscent to those used with the logistic equation, studied as a dynamical system \citep{Devaney-book1986}}
, as shown in the right plot of Fig.~\ref{fig:stability-250}.
Because the updates form a line with a slope whose magnitude is less than 1, it will lead to updates converging to $\rho_{targ}$
\footnote{If desired, further analysis might include a computation of the contraction coefficient about the fixed point $(\rho_{targ},\rho_{targ})$ \citep[see Sec. 8 in][]{Kolmogorov-book1970}.}.
Note that $\rho = \rho_{targ} = 0.1$ is a fixed point in this iterative map.

\begin{figure}[h!]
\begin{center}
\includegraphics[scale=0.32]{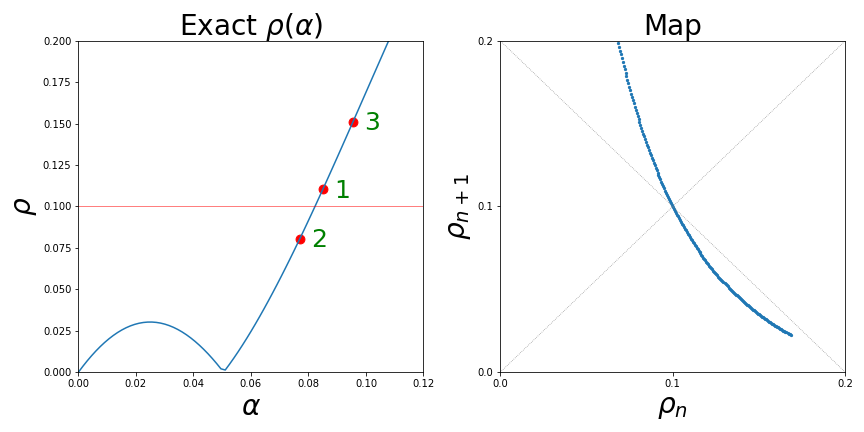}
\captionof{figure}{\color{Green} Here, with $\theta_0 = -2.04$, the iterates lead to divergence away from $\rho_{targ}$, indicated by the red line.  
In the left plot this is shown as a sequence of numbered points which are updates using the AF.
This is reflected in the $\rho$-map in the right plot, where a slope of magnitude greater than 1 indicates unstable updates; this plot was produced using $\alpha \in (0.075,0.095)$.
The gray lines at $\pm 45^\circ$ are added as a visual guide.}
\label{fig:stability-204}
\end{center}
\end{figure}

A second example is made using the starting point $\theta_0 = -2.04$, which is close to the inflection point near $-2.0$.
The left plot in Fig.~\ref{fig:stability-204} reveals more structure, and generally a more nonlinear behavior.
The points labeled 1,2,3 were generated by starting at $\alpha = 0.085$, and repeatedly applying the AF, as before.
They clearly show an unstable behavior in the vicinity of where the $\rho(\alpha)$ curve crosses $\rho = \rho_{targ}$; in this case, $d_n < d_{n+1}$.
Again, the stability can be studied via a $\rho$-map in the right plot; it confirms the unstable behavior,
since the magnitude of the slope of the curve at $\rho = 0.1$ is greater than 1.0.
In applications, inflection points weren't seen by the author to be a significant, practical impediment for this general approach. 
Their effect can be ameliorated by a repeated calculation of $\rho$ and $\alpha$ \citep[see Section 6.5 in][]{neograd_arxiv_v2}.




\newpage

\bibliography{MZproj1}

\begin{thebibliography}{29}
\providecommand{\natexlab}[1]{#1}
\providecommand{\url}[1]{\texttt{#1}}
\expandafter\ifx\csname urlstyle\endcsname\relax
  \providecommand{\doi}[1]{doi: #1}\else
  \providecommand{\doi}{doi: \begingroup \urlstyle{rm}\Url}\fi

\bibitem[Armijo(1966)]{Armijo1966}
L.~Armijo.
\newblock Minimization of functions having {L}ipschitz continuous first partial
  derivatives.
\newblock \emph{Pacific J. Math.}, 16\penalty0 (1):\penalty0 1--3, 1966.

\bibitem[Barzilai and Borwein(1988)]{Barzilai1988}
J.~Barzilai and J.M. Borwein.
\newblock Two-point step size gradient methods.
\newblock \emph{IMA Journal of Numerical Analysis}, 8\penalty0 (1):\penalty0
  141--148, 1988.

\bibitem[Bishop(2006)]{Bishop-book2006}
C.M. Bishop.
\newblock \emph{Pattern Recognition and Machine Learning}.
\newblock Springer, 2006.

\bibitem[Cauchy(1847)]{Cauchy1847}
A.~Cauchy.
\newblock M\'ethode g\'en\'erale pour la r\'esolution des syst\`emes
  d'\'equations simultan\'ees.
\newblock \emph{C. R. Acad. Sci. Paris}, 25:\penalty0 536--538, 1847.

\bibitem[Devaney(1986)]{Devaney-book1986}
R.L. Devaney.
\newblock \emph{An Introduction to Chaotic Dynamical Systems}.
\newblock Benjamin/Cummings, 1986.

\bibitem[Dozat(2016)]{Dozat2016}
T.~Dozat.
\newblock Incorporating {N}esterov momentum into {A}dam.
\newblock \emph{ICLR Workshop}, 1:\penalty0 2013--206, 2016.

\bibitem[Dua and Graff(2017)]{Dua2019}
D.~Dua and C.~Graff.
\newblock {UCI} machine learning repository, 2017.
\newblock URL \url{http://archive.ics.uci.edu/ml}.

\bibitem[Duchi et~al.(2011)Duchi, Hazan, and Singer]{Duchi2011}
J.~Duchi, E.~Hazan, and Y.~Singer.
\newblock Adaptive subgradient methods for online learning and stochastic
  optimization.
\newblock \emph{Journal of Machine Learning Research (JMLR)}, 12:\penalty0
  2121--2159, 2011.

\bibitem[Kingma and Ba(2015)]{Kingma2015}
D.P. Kingma and J.L. Ba.
\newblock Adam: A method for stochastic optimization.
\newblock In \emph{International Conference on Learning Representations
  (ICLR)}, pages 1--13, 2015.

\bibitem[Kolmogorov and Fomin(1970)]{Kolmogorov-book1970}
A.A. Kolmogorov and S.V. Fomin.
\newblock \emph{Introductory Real Analysis}.
\newblock Dover, 1970.

\bibitem[Lemar\'echal(2010)]{Lemarechal2010}
C.~Lemar\'echal.
\newblock Cauchy and the gradient method, 2010.
\newblock URL
  \url{math.uni-bielefeld.de/documents/vol-ismp/40\_lemarechal-claude.pdf}.

\bibitem[Li et~al.(2020)]{Li2020}
F.-F. Li et~al.
\newblock Cs231n: Convolutional neural networks for visual recognition, 2020.
\newblock URL \url{http://cs231n.stanford.edu}.

\bibitem[Nesterov(1983)]{Nesterov1983}
Y.~Nesterov.
\newblock A method for unconstrained convex minimization problem with the rate
  of convergence {O}$(1/k^2)$.
\newblock \emph{Soviet Mathematics Doklady}, 27:\penalty0 372--376, 1983.

\bibitem[Nesterov(2018)]{Nesterov-book2018}
Y.~Nesterov.
\newblock \emph{Lectures on Convex Optimization}.
\newblock Springer, 2nd edition, 2018.

\bibitem[Nocedal and Wright(2006)]{Nocedal-book2006}
J.~Nocedal and S.J. Wright.
\newblock \emph{Numerical Optimization}.
\newblock Springer, 2nd edition, 2006.

\bibitem[Pascanu et~al.(2013)Pascanu, Mikolov, and Bengio]{pascanu2013}
R.~Pascanu, T.~Mikolov, and Y.~Bengio.
\newblock On the difficulty of training neural networks.
\newblock In \emph{International Conference on Machine Learning (ICML)}, 2013.

\bibitem[Polyak(1964)]{Polyak1964}
B.T. Polyak.
\newblock Some methods of speeding up the convergence of iteration methods.
\newblock \emph{USSR Computational Mathematics and Mathematical Physics},
  4\penalty0 (5):\penalty0 1--17, 1964.

\bibitem[Press et~al.(2007)Press, Teukolsky, Vetterling, and
  Flannery]{Press-book2007}
W.H. Press, S.A. Teukolsky, W.T. Vetterling, and B.P. Flannery.
\newblock \emph{Numerical Recipes, The Art of Scientific Computing}.
\newblock Cambridge University Press, 3rd edition, 2007.

\bibitem[Reddi et~al.(2018)Reddi, Kale, and Kumar]{Reddi2018}
S.J. Reddi, S.~Kale, and S.~Kumar.
\newblock On the convergence of {A}dam and beyond.
\newblock In \emph{International Conference on Learning Representations
  (ICLR)}, 2018.

\bibitem[Ruder(2017)]{Ruder2017}
S.~Ruder.
\newblock An overview of gradient descent optimization algorithms.
\newblock \emph{arXiv preprint}, arXiv:1609.04747, 2017.

\bibitem[Scikit-learn(2020)]{scikit-digits}
Scikit-learn.
\newblock Digits dataset, 2020.
\newblock URL
  \url{https://scikit-learn.org/stable/modules/generated/sklearn.datasets.load\_digits.html}.

\bibitem[Sra et~al.(2011)Sra, Nowozin, and Wright]{Sra-book2011}
S.~Sra, S.~Nowozin, and S.J. Wright.
\newblock \emph{Optimization for Machine Learning}.
\newblock The MIT Press, 2011.

\bibitem[Tieleman and Hinton(2012)]{Tieleman2012}
T.~Tieleman and G.~Hinton.
\newblock Lecture 6.5 - {RMSP}rop, {C}oursera: Neural networks for machine
  learning.
\newblock Technical report, 2012.

\bibitem[Wilson et~al.(2017)Wilson, Roelofs, Stern, Srebro, and
  Recht]{Wilson2017}
A.C. Wilson, R.~Roelofs, M.~Stern, N.~Srebro, and B.~Recht.
\newblock The marginal value of adaptive gradient methods in machine learning.
\newblock \emph{arXiv preprint}, arXiv:1705.08292, 2017.

\bibitem[Wolfe(1969)]{Wolfe1969}
P.~Wolfe.
\newblock Convergence conditions for ascent methods.
\newblock \emph{SIAM Review}, 11\penalty0 (2):\penalty0 226–235, 1969.

\bibitem[Wolfe(1971)]{Wolfe1971}
P.~Wolfe.
\newblock Convergence conditions for ascent methods. {II}: Some corrections.
\newblock \emph{SIAM Review}, 13\penalty0 (2):\penalty0 185--188, 1971.

\bibitem[Zeiler(2012)]{Zeiler2012}
M.D. Zeiler.
\newblock Adadelta: an adaptive learning rate method.
\newblock \emph{arXiv preprint}, arXiv:1212.5701, 2012.

\bibitem[Zimmer(2020{\natexlab{a}})]{MZGithub}
M.F. Zimmer.
\newblock Github repository, 2020{\natexlab{a}}.
\newblock URL \url{www.github.com/mfzimmer}.

\bibitem[Zimmer(2020{\natexlab{b}})]{neograd_arxiv_v2}
M.F. Zimmer.
\newblock Neograd: Gradient descent with a near-ideal learning rate, Oct
  2020{\natexlab{b}}.
\newblock URL \url{https://www.arxiv.org/pdf/2010.07873v2.pdf}.

\end{thebibliography}

\end{document}